\pdfoutput=1
\documentclass[11pt]{article}

\usepackage{emnlp2021}

\usepackage{times}
\usepackage{latexsym}
\usepackage{booktabs}

\usepackage[T1]{fontenc}
\usepackage[utf8]{inputenc}

\usepackage{microtype}

\usepackage{graphicx}
\usepackage{multirow}
\usepackage{amsmath}
\usepackage{amsfonts}
\usepackage{tabularx}
\usepackage{color}
\usepackage{comment}
\usepackage{amsmath,amsfonts,amssymb}
\usepackage{arydshln}
\title{Exploring Task Difficulty for Few-Shot Relation Extraction}

\author{
	Jiale Han$^1$, Bo Cheng$^1$ \and Wei Lu$^2$
	\\
	\textsuperscript{1}State Key Laboratory of Networking and Switching Technology,\\Beijing University of Posts and Telecommunications\\
	\textsuperscript{2}StatNLP Research Group, Singapore University of Technology and Design\\
	\texttt{\{hanjl,chengbo\}@bupt.edu.cn, luwei@sutd.edu.sg}
}

\begin{document}
	\maketitle
	\begin{abstract}
		Few-shot relation extraction (FSRE) focuses on recognizing novel relations by learning with merely a handful of annotated instances. Meta-learning has been widely adopted for such a task, which trains on randomly generated few-shot tasks to learn generic data representations. Despite impressive results achieved, existing models still perform suboptimally when handling hard FSRE tasks, where the relations are fine-grained and similar to each other. We argue this is largely because existing models do not distinguish hard tasks from easy ones in the learning process. In this paper, we introduce a novel approach based on contrastive learning that learns better representations by exploiting relation label information. We further design a method that allows the model to adaptively learn how to focus on hard tasks. Experiments on two standard datasets demonstrate the effectiveness of our method.
	\end{abstract}
	
	\section{Introduction}
	{\let\thefootnote\relax\footnote{Accepted as a long paper in EMNLP 2021 (Conference on Empirical Methods in Natural Language Processing).}}
	Relation extraction aims to detect the relation between two entities contained in a sentence, which is the cornerstone of various natural language processing (NLP) applications, including knowledge base enrichment \cite{trisedya-etal-2019-neural}, biomedical knowledge discovery \cite{DBLP:conf/ijcai/GuoN0C20}, and question answering \cite{DBLP:conf/ijcai/HanCW20}. Conventional neural methods \cite{miwa-bansal-2016-end, tran-etal-2019-relation} train a deep network through a large amount of labeled data with extensive relations, so that the model can recognize these relations during the test phase. Although impressive performance has been achieved, these methods are difficult to adapt to novel relations that have never been seen in the training process.  In contrast, humans can identify new relations with very few examples. It is thus of great interest to enable the model to generalize to new relations with a handful of labeled instances. 
	
	\begin{figure}[t!]
		\flushleft
		%\centering
		\includegraphics[width=\linewidth]{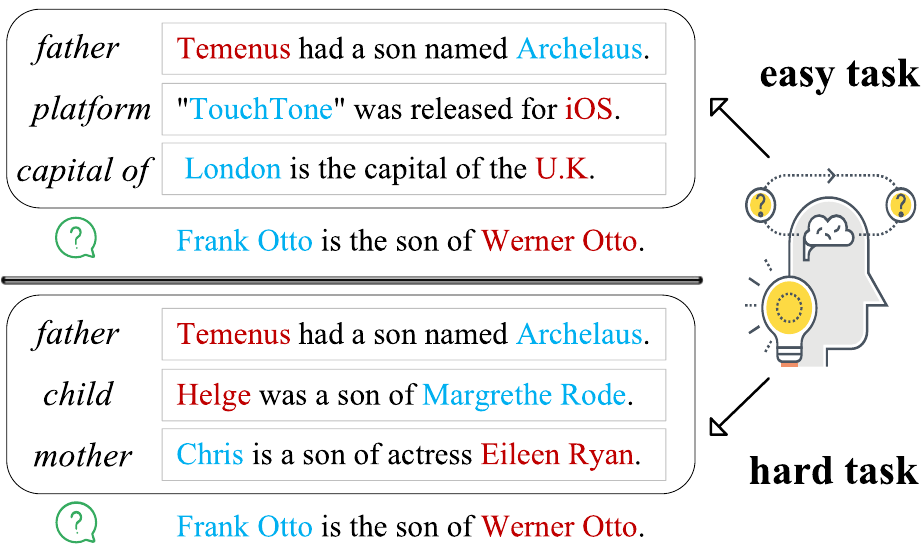}
		\caption{An example of easy few-shot task (top) and hard few-shot task (bottom).
			This is a 3-way-1-shot setup -- each task involves three relations, and each relation has one supporting instance.
			Blue and red colors indicate head and tail entities respectively.
			For the easy task, the relations are very different, and it is easy to classify the query instance. However, due to the subtle differences among the relations in the hard tasks, it is challenging to correctly predict the true relation.}
		\label{intro} 
	\end{figure}
	
	Inspired by the success of few-shot learning in the computer vision (CV) community \cite{DBLP:conf/cvpr/SungYZXTH18, DBLP:conf/iclr/SatorrasE18}, \citet{han-etal-2018-fewrel} first introduce the task of few-shot relation extraction (FSRE). {\color{black}FSRE requires models to be capable of handling classification of novel relations with scarce labeled instances. A popular framework for few-shot learning is meta-learning \cite{DBLP:conf/icml/SantoroBBWL16, DBLP:conf/nips/VinyalsBLKW16}, which optimizes the model through collections of few-shot tasks sampled from the external data containing disjoint relations with novel relations, so that the model can learn cross-task knowledge and use the knowledge to acclimate rapidly to new tasks.} A simple yet effective algorithm based on meta-learning is prototypical network \citep{DBLP:conf/nips/SnellSZ17}, aiming to learn a metric space in which a query instance is classified according to its distance to class prototypes. Recently, many works \cite{DBLP:conf/aaai/GaoH0S19, DBLP:conf/icml/QuGXT20, DBLP:conf/cikm/YangZDHHC20} for FSRE are in line with prototypical networks, which achieve remarkable performance. 
	Nonetheless, the difficulty of distinguishing relations varies in different tasks \cite{DBLP:journals/corr/abs-2007-06240}, depending on the similarity between relations. As illustrated in Figure~\ref{intro}, there are easy few-shot tasks whose relations are quite different, so that they can be consistently well-classified, and also hard few-shot tasks with subtle inter-relation variations which are prone to misclassification. 
	Current FSRE methods struggle with handling the hard tasks given limited labeled instances due to two main reasons. First, most works mainly focus on general tasks to learn generalized representations, and ignore modeling subtle and local differences of relations effectively, which may hinder these models from dealing with hard tasks well. Second, current meta-learning methods treat training tasks equally, which are randomly sampled and have different degrees of difficulty. The generated easy tasks can overwhelm the training process training and lead to a degenerate model.
	
	{\color{black}To fill this gap, this paper proposes a {\em {H}ybrid {C}ontrastive {R}elation-{P}rototype} (HCRP) approach, which focuses on improving the performance on hard FSRE tasks.} 
	{\color{black}Concretely, we first propose a hybrid prototypical network, capable of capturing global and local features to generate the informative class prototypes.}
	Next, we present a novel relation-prototype contrastive learning method, which leverages relation descriptions as anchors, and pulls the prototype of the same class closer in representation space and pushes those of different classes away.
	In this way, the model gains diverse and discriminative prototype representations, which could be beneficial to distinguish the subtle difference of confusing relations in hard few-shot tasks. Furthermore, we design a task-adaptive training strategy based on focal loss \citep{DBLP:conf/iccv/LinGGHD17} to learn more from hard tasks, which allocates dynamic weights to different tasks according to task difficulty. Extensive experiments on two large-scale benchmarks show that our model significantly outperforms the baselines. Ablation and case studies demonstrate the effectiveness of the proposed modules. Our code is available at {\url{https://github.com/hanjiale/HCRP}} .
	
	The contributions of this paper are summarized as follows:
	\begin{itemize}
		\item {\color{black}We present HCRP to explore task difficulty as useful information for FSRE, which boosts hybrid prototypical network with relation-prototype contrastive learning to capture diverse and discriminative representations.}
		\item We design a novel task adaptive focal loss to focus training on hard tasks, which enables the model to achieve higher robustness and better performance.
		
		\item Qualitative and quantitative experiments on two FSRE benchmarks demonstrate the effectiveness of our model.
	\end{itemize}
	
	\section{Related Work}
	
	\subsection{Few-shot Relation Extraction}
	Relation extraction is a foundational and important task in NLP and attracts many recent attentions \cite{nan-etal-2020-reasoning, DBLP:journals/corr/abs-2104-07650}. Few-shot relation extraction aims to predict novel relations by exploring a few labeled instances. \citet{han-etal-2018-fewrel} first present a large-scale benchmark FewRel for FSRE.
	\citet{DBLP:conf/aaai/GaoH0S19} design a hybrid attention-based prototypical network to highlight the crucial instances and features. \citet{ye-ling-2019-multi} propose a prototypical network with multi-level matching and aggregation. 
	\citet{sun-etal-2019-hierarchical} present a hierarchical attention prototypical network to enhance the representation ability of semantic space. 
	\citet{DBLP:conf/icml/QuGXT20} utilize an external relation graph to study the relationships between different relations. \citet{wang-etal-2020-learning} apply added relative position information and syntactic relation information to enhance prototypical networks. \citet{DBLP:conf/cikm/YangZDHHC20} fuse text descriptions of relations and entities by a collaborative attention mechanism. 
	And \citet{yang-etal-2021-entity} introduce the inherent concepts of entities to provide clues for relation classification. There are also some methods \cite{baldini-soares-etal-2019-matching, peng-etal-2020-learning} combining prototypical networks with pre-trained language models, which achieve impressive results. {\color{black}However, the task difficulty of FSRE has not been explored. In this work, we focus on the hard tasks and propose a hybrid contrastive relation-prototype method to better model subtle {\color{black}variations across different relations}.}
	
	\subsection{Contrastive Learning}
	
	\begin{figure*}[tbp!]
		%\flushleft
		\centering 
		\includegraphics[width=0.8\textwidth]{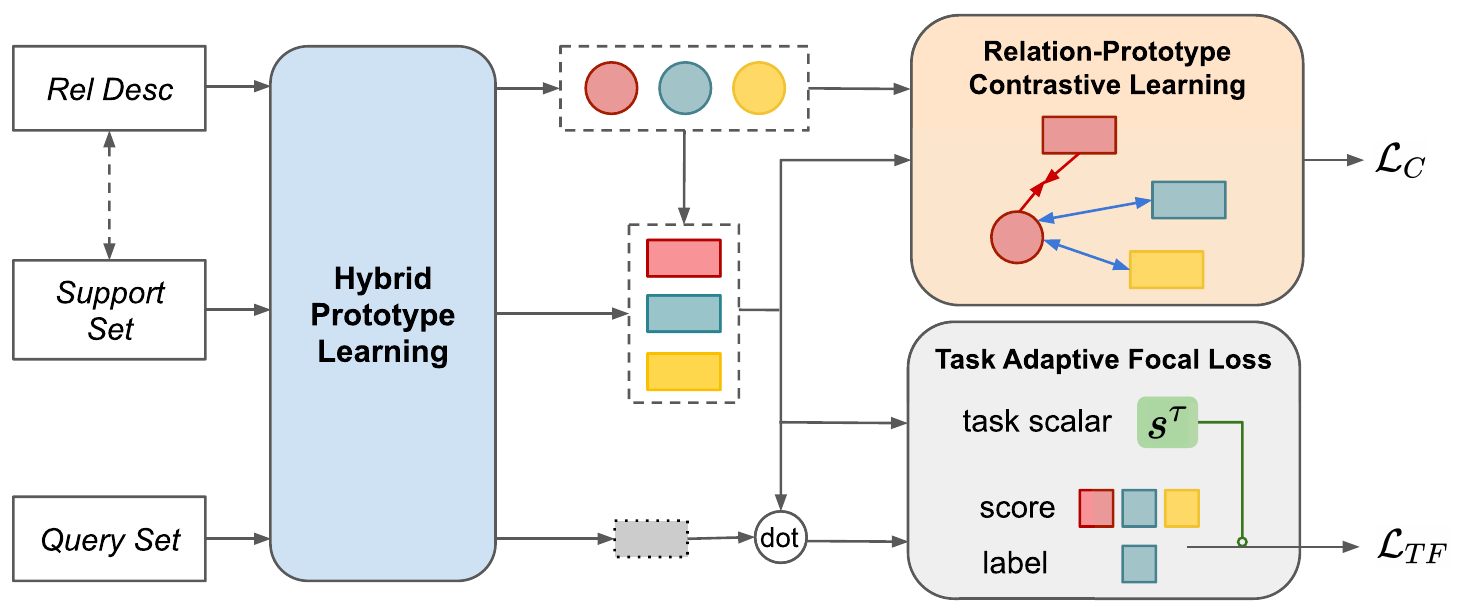}
		\caption{The overall framework of HCRP. Best viewed in color. The rectangles represent the class prototypes, the circles represent the relations, and different colors represent different classes.
		}
		\label{framework} 
	\end{figure*}
	Contrastive learning \cite{DBLP:journals/corr/abs-2011-00362} has gained popularity recently in the CV community. The core idea is to contrast the similarities and dissimilarities between data instances, which pulls the positives closer and pushes negatives away simultaneously. CPC \cite{DBLP:journals/corr/abs-1807-03748} proposes a universal unsupervised learning approach. MoCo \cite{DBLP:conf/cvpr/He0WXG20} presents a mechanism for building dynamic dictionaries for contrastive learning. SimCLR \cite{DBLP:conf/icml/ChenK0H20} improves contrastive learning by using larger batch size and data augmentation. \citet{DBLP:conf/nips/KhoslaTWSTIMLK20} extend the self-supervised contrastive approach to the supervised setting.
	\citet{Nan_2021_CVPR} propose a dual contrastive learning approach for video grounding. There are also some applications of contrastive learning in the field of NLP. \citet{DBLP:journals/corr/abs-2005-12766} employ back translation and MoCo to learn sentence-level representations. \citet{gunel2021supervised} design supervised contrastive learning for pre-trained language model fine-tuning. Inspired by these works, we propose a heterogeneous relation-prototype contrastive learning in a supervised way to obtain more discriminative representations.
	
	\section{Task Definition}
	
	{\color{black}We follow a typical few-shot task setting, namely the $N$-way-$K$-shot setup}, which contains a support set $\mathcal{S}$ and a query set $\mathcal{Q}$. The support set $\mathcal{S}$ includes $N$ novel classes, each with $K$ labeled instances. The query set $\mathcal{Q}$ contains the same $N$ classes as $\mathcal{S}$. And the task is evaluated on query set $\mathcal{Q}$, trying to predict the relations of instances in $\mathcal{Q}$. What's more, an auxiliary dataset $\mathcal{D}_{base}$ is given, which contains abundant base classes, each with a large number of labeled examples. Note the base classes and novel classes are disjoint with each other. 
	The few-shot learner aims to acquire knowledge from base classes and use the knowledge to recognize novel classes. 
	One popular approach is the meta-learning paradigm \cite{DBLP:conf/nips/VinyalsBLKW16}, which mimics the few-shot learning settings at training stage. Specifically, in each training iteration, we randomly select $N$ classes from base classes, each with $K$ instances to form a support set $\mathcal{S}=\{s_k^i; i=1, \dots , N, k=1, \dots , K\}$. Meanwhile, $R$ instances are sampled from the remaining data of the $N$ classes to construct a query set $\mathcal{Q}=\{q_j; j=1, \dots , R\}$. The model is optimized by collections of few-shot tasks sampled from base classes, so that it can rapidly adapt to new tasks.
	
	For an FSRE task, each instance consists of a set of samples $(x, e, y)$, where $x$ denotes a natural language sentence, $e=(e_{h}, e_{t})$ indicates a pair of head entity and tail entity, and $y$ is the relation label. The name and description for each relation are also provided as auxiliary support evidence for relation extraction. For example, for a relation with its relation id ``P726'' in a dataset that we use, we can obtain its name ``\textit{candidate}'' and description ``\textit{person or party that is an option for an office in an election}''. 
	
	\section{Approach}
	In this section, we present the details of our proposed HCRP approach.
	The overall learning framework is illustrated in Figure~\ref{framework}.
	The inputs are $N$-way-$K$-shot tasks (sampled from the auxiliary dataset $\mathcal{D}_{base}$), where each task contains a support set $\mathcal{S}$ and a query set $\mathcal{Q}$.
	Meanwhile, we take the names and descriptions of these $N$ classes (i.e., relations) as inputs as well.
	{\color{black}HCRP consists of three components. The hybrid prototype learning module generates informative prototypes by capturing global and local features, which can better capture the subtle differences of relations. The relation-prototype contrastive learning component is then used to leverage the relation label information to further enhance the discriminative power of the prototype representations. Finally, a task adaptive focal loss is introduced to encourage the model to focus training on hard tasks.}
	
	\subsection{Hybrid Prototype Learning}\label{4.1}
	%	\subsection{Relational Prototype Learning}\label{4.1}
	We employ BERT \cite{devlin-etal-2019-bert} as the encoder to obtain contextualized embeddings of query instances $\{\mathbf{Q}_j \in\mathbb{R}^{l_{q_j}\times d}; j=1, \dots , R\}$ and support instances $\{\mathbf{S}_k^i \in\mathbb{R}^{l_{s_k^i}\times d}; i=1, \dots , N, k=1, \dots , K\}$, {\color{black}where $l_{q_j}$ and $l_{s_k^i}$ are the sentence lengths of the $j$-th query instance and $k$-th support instance in class $i$ respectively}, and $d$ is the size of the resulting contextualized representations. For each relation, we concatenate the name and description and feed the sequence into the BERT encoder to obtain relation embeddings $\{\mathbf{R}^i \in\mathbb{R}^{l_{r^i}\times d}; i=1, \dots , N\}$, {\color{black}where $l_{r^i}$ is the length of relation description $i$}.
	\subsubsection*{Global Prototypes}
	
	%	\paragraph{Global Features.}
	For instances in $\mathcal{S}$ and $\mathcal{Q}$, the global features $\{\mathbf{s}_k^i \in\mathbb{R}^{2d}; i=1, \dots , N, k=1, \dots , K\}$ and $\{\mathbf{q}_j \in\mathbb{R}^{2d}; j=1, \dots , R\}$ are obtained by concatenating the hidden states corresponding to start tokens of two entity mentions following \citet{baldini-soares-etal-2019-matching}.
	The global features of relations $\{\mathbf{r}^i \in\mathbb{R}^{2d}; i=1, \dots , N\}$ are obtained by the hidden states corresponding to [\textit{CLS}] token (converted to $2d$ dimension with a transformation).
	For each relation $i$, we average the global features of the $K$ supporting instances {\color{black}following the work of \citet{DBLP:conf/nips/SnellSZ17}}, and further add the global feature of relation to form global prototype representation.
	\begin{equation}
		\mathbf{p}_g^i = \frac{1}{K}\sum_{k=1}^{K} \mathbf{s}_k^i + \mathbf{r}^i \in \mathbb{R}^{2d}
	\end{equation}
	
	\subsubsection*{Local Prototypes}
	
	%	\paragraph{Local Features.}
	%To better handle the fine-grained FSRE task, we further capture local representation to highlight the focus part. Specifically, the model applies attention mechanism to get the local features of support set and relation text. For a relation $n$, the local feature of support instance is calculated as:
	{\color{black}While global prototypes are capable of capturing general data representations, such representations may not readily capture useful local information within specific RSRE tasks. 
		To better handle the hard FSRE tasks with subtle  differences among highly similar relations, we further propose local prototypes to highlight key tokens in an instance that are essential to {\color{black}characterize different relations}.}

	For relation $i$, we first calculate the local feature of the $k$-th support instance as:
	\begin{eqnarray}
		\hat{\mathbf{s}}_k^i &=& \sum_{n=1}^{l_{s_k^i}}\alpha^s_n {[\mathbf{S}_k^i]}_n  \in \mathbb{R}^{d}
		\\
		\alpha^s &=& {\rm softmax}({\rm sum}(\mathbf{S}_k^i{(\mathbf{R}^i)}^\mathsf{T}))  \in \mathbb{R}^{l_{s_k^i}}
	\end{eqnarray}
	where ${[\cdot]}_n$ is the $n$-th row of a matrix, ${\rm sum}()$ is an operation that sums all elements for each row in a matrix. 
	\begin{comment}
	{\color{purple}LW: need to add a few sentences to explain what's going on here. Also, explain why this can be interpreted as an attention mechanism, if you would like to call it that way (you used self-attention below).}
	\end{comment}
	{\color{black}Specifically, we allocate weights to different tokens according to their similarities with relation descriptions, and take the weighted sum to form such local features.}
	
	{\color{black} Similarly, we calculate the similarity between relation embedding $\mathbf{R}^i$ and each support instance embedding $\mathbf{S}_k^i$ of relation $i$ and obtain $K$ features
		$\{\hat{\mathbf{r}}_k^i; k=1, \dots , K\}$:
		%\begin{comment}
		\begin{eqnarray}
			\hat{\mathbf{r}}_k^i &=& \sum_{n=1}^{l_{r^i}}\alpha^r_n[\mathbf{R}^i]_{n} \in \mathbb{R}^{d}
			\\
			\alpha^r &=& {\rm softmax}({\rm sum}(\mathbf{R}^i{(\mathbf{S}_k^n)}^\mathsf{T}))  \in \mathbb{R}^{l_{r^i}}
		\end{eqnarray}
		%	\end{comment}
		
		The $K$ features are then averaged to arrive at the final local representation of relation $i$:}
	\begin{equation}
		\hat{\mathbf{r}}^i = \frac{1}{K}\sum_{k=1}^{K}\hat{\mathbf{r}}_k^i  \in \mathbb{R}^{d}
	\end{equation}
	
	The local feature of a query instance is calculated by the following formulas.
	\begin{eqnarray}
		\hat{\mathbf{q}}_j &=& \sum_{n=1}^{l_{q_j}}\alpha^q_n {[\mathbf{Q}_j]}_n  \in \mathbb{R}^{d}
		\\
		\alpha^q &=& {\rm softmax}({\rm sum}(\mathbf{Q}_j\mathbf{Q}_j^\mathsf{T}))  \in \mathbb{R}^{l_{q_j}}
	\end{eqnarray}
	
	Finally, we generate the local prototype by averaging the local features of the support set, plus the local feature of the relation.
	\begin{equation}
		\mathbf{p}_l^i = \frac{1}{K}\sum_{k=1}^{K} \hat{\mathbf{s}}_k^i + \hat{\mathbf{r}}^i \in \mathbb{R}^{d}
	\end{equation}
	
	\subsubsection*{Hybrid Prototypes}
	%	With the obtained global and local features, we propose the hybrid prototype module to generate diverse and pivotal prototype representation.
	%	\paragraph{Local Prototype.} 
	%	\paragraph{Hybrid Prototype.}
	The model concatenates the global and local prototype to form hybrid prototype representations:
	\begin{equation}
		\mathbf{p}_h^i = [\mathbf{p}_g^i; \mathbf{p}_l^i]\in \mathbb{R}^{3d}
	\end{equation}
	where $[;]$ denotes column-wise concatenation. The hybrid representation of query instance is also obtained by concatenating the global and local features:
	\begin{equation}
		\mathbf{q}^j_h = [\mathbf{q}_j; \hat{\mathbf{q}}_j]\in \mathbb{R}^{3d}
	\end{equation}
	
	With the representation of query and prototypes of $N$ relations, the model computes the probability of the relations for the query instance $q_j$ as follows:
	\begin{equation}
		z(y=i|q_j) = \frac{\mathrm{exp}(\mathbf{q}^j_h \cdot \mathbf{p}_h^i)}{\sum_{n=1}^{N}\mathrm{exp}(\mathbf{q}^j_h \cdot \mathbf{p}_h^n)}
	\end{equation}
	
	\subsection{Relation-Prototype Contrastive Learning}\label{4.3}
	
	{\color{black}Hard tasks usually involve similar relations whose prototype representations are close, leading to increased challenges in classifying query instances.}
	To gain a more discriminative prototype representation, we design a novel Relation-Prototype Contrastive Learning (RPCL) method, which leverages the interpretable relation names and descriptions to calibrate the few-shot prototypes. Unlike conventional unsupervised or self-supervised contrastive learning, RPCL utilizes the labels of support instances in each task to {\color{black}perform supervised contrastive learning}.
	
	Concretely, taking a relation representation as an anchor, the prototype of the same class as positive and prototypes of different classes as negatives, RPCL aims to pull the positive closer with the anchor and pushes negatives away. For a specific relation $i$ with its hybrid representation,
	\begin{equation}
		\mathbf{r}^i_h = [\mathbf{r}^i;\hat{\mathbf{r}}^i]\in \mathbb{R}^{3d}
	\end{equation}
	the model collects positive prototype $\mathbf{p}_h^i$ and negative prototypes $\{\mathbf{p}_h^n; n=1, \dots ,N, n\neq i\}$.  {\color{black}The goal is to distinguish the positive from the negatives}. We use dot product to measure the similarities between the relation anchor and selected prototypes.
	\begin{eqnarray}
		u^i_{pos} &=& \mathbf{p}_h^i \cdot  \mathbf{r}^i_h \in \mathbb{R}
		\\
		u^{i,n}_{neg} &=& \mathbf{p}_h^n  \cdot \mathbf{r}^i_h \in \mathbb{R}
	\end{eqnarray}
	%where $\mathbf{P}_h^i \in \mathbb{R}^{(N-1) \times 3d}$ represents the matrix of negative prototypes.  
	The contrastive loss is calculated by the following formula:
	\begin{comment}
	\begin{align}
	\mathcal{L}_C &= \sum_{i=1}^{N}\mathcal{L}_c^i \\
	&=\sum_{i=1}^{N}-\mathrm{log} \frac{u^i_{pos}}{u^i_{pos} + {\rm sum}(u^i_{neg})}
	\end{align}
	\end{comment}
	\begin{align}
		\mathcal{L}_C
		&=\sum_{i=1}^{N}-\mathrm{log} \frac{u^i_{pos}}{u^i_{pos} + \sum_{n}u^{i,n}_{neg}}
	\end{align}
	
	\subsection{Task Adaptive Focal Loss}\label{4.4}
	%In order to avoid model degradation due to simple coarse-grained tasks, 
	We design a task adaptive focal loss to learn more from hard tasks, which is a modified cross entropy (CE) loss.
	The CE loss can be written as follows:
	\begin{equation}
		\mathcal{L}_{CE}=-\mathrm{log}(z_y)
	\end{equation}
	where $y$ is the class label, and $z_y$ is the estimated probability for the class $y$. The focal loss proposed by \citet{DBLP:conf/iccv/LinGGHD17} aims to solve the imbalance of hard examples and easy examples.
	\begin{equation}
		\mathcal{L}_{F}=-(1-z_y)^\gamma \mathrm{log}(z_y)
	\end{equation}
	where $\gamma \geq 0$ adjusts the rate at which easy examples are down-weighted. 
	For an easy example, $z_y$ is almost 1, the factor goes to 0, and the loss for easy examples is down-weighted, which in turn increases the importance of correcting misclassified examples, {\color{black}which are potentially harder}.
	
	We employ focal loss instead of cross entropy loss to focus more on hard {query examples}.
	Moreover, to focus more on hard tasks, we design a novel {\em task adaptive focal loss}, which introduces the dynamic task-level weights. Specifically, for an $N$-way-$K$-shot task, the model calculates the class-wise similarity to estimate task difficulty. The higher the inter-class similarity, the harder the task. We first concatenate the hybrid features of prototype and relation to represent each class $\mathbf{c}^i=[\mathbf{r}^i_h;\mathbf{p}_h^i]$,
	\begin{comment}
	\begin{equation}
	\mathbf{c}^i=[\mathbf{r}^i_h;\mathbf{p}_h^i]\in \mathbb{R}^{6d}
	\end{equation}
	\end{comment}
	and then define the task similarity matrix $\mathbf{S}^{\tau}\in \mathbb{R}^{N\times N}$, for $i, j \in \{1, \dots , N\}$, 
	\begin{equation}
		\mathbf{s}^{\tau}_{ij}=\frac{\mathbf{c}^i \cdot \mathbf{c}^j}{||\mathbf{c}^i|| \times ||\mathbf{c}^j||}
	\end{equation}
	where $||\cdot||$ is the Euclidean norm. The task similarity scalar is obtained by the following formula:
	\begin{equation}
		s^{\tau}=\frac{\mathrm{exp}(||\mathbf{S}^{\tau}||_\mathrm{F})}{\sum_{\tau^{'}=1}^{\mathcal{T}}\mathrm{exp}(||\mathbf{S}^{\tau^{'}}||_\mathrm{F})}
	\end{equation}
	where $||\cdot||_\mathrm{F}$ is the Frobenius norm, and $\mathcal{T}$ is the {\color{black}number of tasks} in a mini-batch. The scalar represents the  degree of difficulty of the task. We add the task-level scalar to the focal loss, which not only focuses on the hard examples at the instance level, but also focuses more on the hard tasks at the task level. Formally, the task adaptive focal loss is defined as follows,
	\begin{equation}
		\mathcal{L}_{TF}=-s^{\tau}(1-z_y)^\gamma \mathrm{log}(z_y)
	\end{equation}
	
	The final objective function of our model is defined as $\mathcal{L}=\mathcal{L}_{TF} + \lambda\times\mathcal{L}_{C} $, where $\lambda$ is a hyper-parameter to balance the two terms.
	\section{Experiments}
	\subsection{Experimental Setup}
	\subsubsection{Datasets}
	We evaluate our model on FewRel 1.0 \cite{han-etal-2018-fewrel} and FewRel 2.0 \cite{gao-etal-2019-fewrel}. FewRel 1.0 and FewRel 2.0 are large-scale few-shot relation extraction datasets, consisting of 100 relations, each with 700 labeled instances. The average number of tokens in each sentence instance is 24.99, and there are 124,577 unique tokens in total. 
	Our experiments follow the splits used in official benchmarks, which split the dataset into 64 base classes for training, 16 classes for validation, and 20 novel classes for testing. FewRel 1.0 is trained and tested on the same Wikipedia domain. In addition, the name and description of each relation are also given, providing additional interpretability for each relation. {\color{black}FewRel 2.0 with domain adaptation setting is trained on Wikipedia domain but tested on a different biomedical domain. Only the names of relation labels are given but descriptions are not available, which makes the task more challenging.}
	\begin{table}[t!]
		\centering
		\scalebox{0.8}{
			\begin{tabular}{ccc}
				\toprule
				Component & Parameter & Value\\
				\midrule
				\multirow{3}*{BERT}
				& type & base-uncased \\
				&hidden size&$768$\\
				%	&layers&12\\
				%	&parameters & 110M \\
				&max length & $128$ \\
				\midrule
				\multirow{3}*{Training}
				& learning rate & $2e-5$  \\
				&batch size& $4$\\
				&max iterations&$30,000$\\
				%	&training N&10\\
				%	&training K&1\\
				\midrule
				\multirow{2}*{Loss}
				&$\lambda$&$1$ / $2.5$\\
				&$\gamma$&1\\
				\bottomrule
			\end{tabular}
		}
		\caption{\label{hyper-parameters}Hyper-parameters (FewRel 1.0 / 2.0) of our approach. %The numbers in parentheses ($\cdot$) represent the hyperparameter values for FewRel 2.0.
		}
	\end{table}
	
	\subsubsection{Evaluation}\label{5.1.2}
	Consistent with the official evaluation scripts, we evaluate our model by randomly sampling 10,000 tasks from validation data. The performance of the model is evaluated as the averaged accuracy on the query set of multiple $N$-way-$K$-shot tasks. According to the previous work \cite{han-etal-2018-fewrel, gao-etal-2019-fewrel}, we choose $N$ to be 5 and 10, and $K$ to be 1 and 5 to form 4 scenarios. {\color{black}We report the final test accuracy by submitting the prediction of our model to the FewRel leaderboard\footnote[2]{\url{https://thunlp.github.io/fewrel.html}}.}
	
	\subsubsection{Implementation Details}
	
	The approach is implemented with PyTorch \cite{DBLP:conf/nips/PaszkeGMLBCKLGA19} and trained on 1 Tesla P40 GPU. We adopt the Transformer library of Huggingface\footnote[3]{\url{https://github.com/huggingface/transformers}}  \cite{wolf-etal-2020-transformers} and take the uncased model of BERT$_{base}$ as the encoder for fair comparison. The AdamW optimizer \cite{DBLP:conf/iclr/LoshchilovH19} is applied to minimize loss. {\color{black}We manually adjust the hyper-parameters based on the performance on the validation data, which are listed in Table~\ref{hyper-parameters}.} Specifically, we use the same hyper-parameter values for two datasets except for $\lambda$. For FewRel 1.0, we concatenate the name and description of each relation as inputs, and $\lambda$ is set to 1. For FewRel 2.0, we only input the relation names, and $\lambda$ is adjusted to 2.5. The number of parameters in our model is 110 million. The average runtime of training and evaluation under 10-way-1-shot setting is 13.35 hours and 1.25 hours, respectively.
	{\color{black}\subsection{Results and Discussion}
		\subsubsection{{\color{black}Comparison to Baselines}}\label{5.2.1}}
	\begin{table*}[th]
		\centering
		\scalebox{0.8}{
			\begin{tabular}{clcccc}
				\toprule
				Encoder & Model&5-way-1-shot&5-way-5-shot&10-way-1-shot&10-way-5-shot\\
				\midrule
				%\textbf{\textit{CNN Encoder}}&&&&\\
				\multirow{3}*{\rotatebox{90}{CNN}} &Proto-CNN$^\clubsuit$ \citep{DBLP:conf/nips/SnellSZ17} &72.65 / 74.52 &86.15 / 88.40 &60.13 / 62.38 &76.20 / 80.45 \\
				&Proto-HATT \citep{DBLP:conf/aaai/GaoH0S19} &75.01 / --- --- & 87.09 / 90.12 &62.48 / --- ---&77.50 / 83.05 \\
				&MLMAN \cite{ye-ling-2019-multi} &79.01 / 82.98 & 88.86 / 92.66 &67.37 / 75.59&80.07 / 87.29 \\
				\midrule
				%	\textbf{\textit{BERT Encoder}}&&&&\\
				\multirow{11}*{\rotatebox{90}{BERT}} &Proto-BERT$^\ast$ \citep{DBLP:conf/nips/SnellSZ17} &82.92 / 80.68 &91.32 / 89.60 &73.24 / 71.48 &83.68 / 82.89 \\
				&MAML$^\ast$ \cite{DBLP:conf/icml/FinnAL17} &82.93 / 89.70 & 86.21 / 93.55 &73.20 / 83.17&76.06 / 88.51 \\
				&GNN$^\ast$ \cite{DBLP:conf/iclr/SatorrasE18} &--- --- / 75.66 & --- --- / 89.06 &--- --- / 70.08&--- --- / 76.93 \\
				&BERT-PAIR$^\clubsuit$ \cite{gao-etal-2019-fewrel} &85.66 / 88.32 & 89.48 / 93.22 &76.84 / 80.63&81.76 / 87.02 \\
				&REGRAB \cite{DBLP:conf/icml/QuGXT20} & 87.95 / 90.30 & 92.54 / 94.25 &80.26 / 84.09&86.72 / 89.93 \\
				&TD-Proto \cite{DBLP:conf/cikm/YangZDHHC20} & --- --- / 84.76 &  --- --- / 92.38 &--- --- / 74.32& --- --- / 85.92 \\
				&CTEG \cite{wang-etal-2020-learning} &84.72 / 88.11 & 92.52 / 95.25 &76.01 / 81.29&84.89 / 91.33 \\
				&\textbf{HCRP (ours)} &\textbf{90.90} / \textbf{93.76}   & \textbf{93.22} /
				\textbf{95.66}   &\textbf{84.11} / \textbf{89.95} & \textbf{87.79} / \textbf{92.10} \\
				\cline{2-6}
				%\hdashline
				%\cdashline
				&MTB$^\star$ \cite{baldini-soares-etal-2019-matching} &--- --- / 91.10 & --- --- / 95.40 &--- --- / 84.30&--- --- / 91.80 \\
				&CP$^\star$ \cite{peng-etal-2020-learning}&--- --- / 95.10 & --- --- / 97.10 &--- --- / 91.20&--- --- / 94.70 \\
				&\textbf{HCPR+CP} & \textbf{94.10} / \textbf{96.42} & \textbf{96.05} / \textbf{97.96}& \textbf{89.13}  /  \textbf{93.97} & \textbf{93.10} /  \textbf{96.46}
				\\
				\bottomrule
			\end{tabular}
		}
		\caption{Accuracy (\%) of few-shot classification on the FewRel 1.0 validation / test set. $\clubsuit$ are from FewRel public leaderboard\footnotemark[2], $\ast$ are reported by \citet{DBLP:conf/icml/QuGXT20}, and $\star$ are reported by \citet{peng-etal-2020-learning}. {\color{black}Our method introduces additional relation label name and description information, which is the same as TD-Proto. Other baseline methods also use different external knowledge. See Section~\ref{5.2.1} for details.}}
		\label{main-fewrel1}
	\end{table*}
	\begin{table}
		\centering
		\renewcommand\tabcolsep{3.6pt}
		\scalebox{0.8}{
			\begin{tabular}{lcccc}
				\toprule
				\multirow{2}*{Model} 
				&5-way&5-way&10-way& 10-way\\
				&1-shot&5-shot&1-shot& 5-shot\\
				\midrule 
				Proto-CNN &35.09 &49.37  &22.98  &35.22  \\
				Proto-BERT   &40.12  & 51.50  &26.45 &36.93  \\
				Proto-ADV  & 42.21 &  58.71 &28.91&44.35 \\
				BERT-PAIR &67.41 & 78.57&54.89&66.85 \\
				HCRP (ours) & \textbf{76.34} &\textbf{83.03}  & \textbf{63.77}&\textbf{72.94} \\
				\bottomrule
			\end{tabular}
		}
		\caption{Accuracy (\%) of few-shot classification on the FewRel 2.0 domain adaptation test set. All results of baselines are quoted from FewRel leaderboard\footnotemark[2].}
		\label{main-fewrel2}
	\end{table}
	We compare our model with the following baseline methods: 1) \textbf{Proto} \cite{DBLP:conf/nips/SnellSZ17}, the algorithm of prototypical networks. We employ CNN and BERT as encoder separately (\textbf{Proto-CNN} and \textbf{Proto-BERT}), and combine adversarial training (\textbf{Proto-ADV}) for FewRel 2.0 domain adaptation. 2) \textbf{MAML} \cite{DBLP:conf/icml/FinnAL17}, the model-agnostic meta-learning algorithm. 3) \textbf{GNN} \cite{DBLP:conf/iclr/SatorrasE18}, a meta-learning approach using graph neural networks. 4) \textbf{Proto-HATT} \cite{DBLP:conf/aaai/GaoH0S19}, prototypical networks modified with hybrid attention to focus on the crucial instances and features. 5) \textbf{MLMAN} \cite{ye-ling-2019-multi}, a multi-level matching and aggregation prototypical network. 6) \textbf{BERT-PAIR} \cite{gao-etal-2019-fewrel}, a method that measures similarity of sentence pair. 7) \textbf{REGRAB} \cite{DBLP:conf/icml/QuGXT20}, a Bayesian meta-learning method with an external global relation graph. 8) \textbf{TD-Proto} \cite{DBLP:conf/cikm/YangZDHHC20}, enhancing prototypical network with both relation and entity descriptions. 9) \textbf{CTEG} \cite{wang-etal-2020-learning}, a model that learns to decouple high co-occurrence relations, where two external information are added. Moreover, we compare our model with two pre-trained RE methods: 10) \textbf{MTB} \cite{baldini-soares-etal-2019-matching}, pre-train with their proposed matching the blank task on top of an existing BERT model. {\color{black}11) \textbf{CP} \cite{peng-etal-2020-learning}, an entity-masked contrastive pre-training framework for RE. %The model constructs positive sentence pairs with the same relation and negative pairs with different relations from Wikidata,  and employs contrastive learning to optimize the network. 
		They first construct a large-scale dataset from Wikidata for pre-training, which contains 744 relations and 867,278 sentences. They then continue pre-training an existing BERT model on such a new dataset, and fine-tune on the FewRel data based on prototypical networks, achieving high accuracy.} Note we do not adopt the results in \citet{baldini-soares-etal-2019-matching} because of their BERT$_{large}$ backbone employed. Here we report the experimental results produced by the work of \citet{peng-etal-2020-learning} which is based on BERT$_{base}$ as the encoder for fair comparison.
	
	Table~\ref{main-fewrel1} presents the experimental results on FewRel 1.0 validation set and test set. 
	As shown in the upper part of Table~\ref{main-fewrel1}, our method outperforms the strong baseline models by a large margin, especially in 1-shot scenarios. Specifically, we improve 5-way-1-shot and 10-way-1-shot tasks 3.46 points and 5.86 points in terms of accuracy respectively compared to the second best method, demonstrating the superior generalization ability. Our method also achieves the best performance on FewRel 2.0, as shown in Table~\ref{main-fewrel2}, which proves the stability and effectiveness of our model. The performance gain mainly comes from three aspects. (1) The hybrid prototypical networks capture rich and subtle features. (2) The relation-prototype contrastive learning leverages the relation text to further gain discriminative prototypes. (3) The task-adaptive focal loss forces model to learn more from hard few-shot tasks. In addition, we evaluate our approach based on the model CP, where the BERT encoder is initialized with their pre-trained parameters\footnote[4]{\url{https://github.com/thunlp/RE-Context-or-Names}}. The lower part of Table~\ref{main-fewrel1} shows that our approach achieves a consistent performance boost when using their pre-trained model, which demonstrates the effectiveness of our method, and also indicates the importance of good representations for few-shot tasks.
	\subsubsection{Performance on Hard Few-shot Tasks}\label{5.3}
	To further illustrate the effectiveness of the developed method, especially for hard FSRE tasks, we evaluate the models on FewRel 1.0 validation set with three different 3-way-1-shot settings, as shown in Table~\ref{fg-fewrel}. \textit{Random} is the general evaluation setting, which samples 10,000 test tasks randomly from validation relations, as detailed in section~\ref{5.1.2}. \textit{Easy} represents the evaluated tasks are easy. We fix the 3 relations in each task as 3 very different relations, which are ``\textit{crosses}'', ``\textit{constellation}'', and ``\textit{military rank}''. Different tasks own different instances but the same relations. Similarly, we pick 3 similar relations, which are ``\textit{mother}'', ``\textit{child}'', and ``\textit{spouse}'' respectively, and evaluate the performance of models under the \textit{Hard} setting. As we can see, the baselines achieve good performance under random and easy settings. However, the accuracy has dropped significantly under the hard setting, which illustrates that hard few-shot tasks are extremely challenging. HCRP gains the best accuracy, especially under the hard setting, proving that our model can effectively handle hard few-shot tasks.
	
	\subsubsection{Analysis of Hybrid Prototype Learning}
	\begin{table}
		\centering
		\renewcommand\tabcolsep{9pt}
		\scalebox{0.8}{
			\begin{tabular}{lccc}
				\toprule
				%\multirow{2}*{Model} &
				%	\multicolumn{2}{c}{fine-grain}\\
				Model &Random &Easy&Hard\\
				\midrule
				Proto-HATT  & 83.96 &  94.62& 35.54 \\
				MLMAN  & 86.37 & 97.30 & 34.45\\
				Proto-BERT   &87.37  & 98.51  &35.63\\
				BERT-PAIR  & 91.14 &99.76  & 38.21\\
				%\hline
				%w/o local &92.98  &99.68 &61.43 \\
				%w/o RPCL & 90.15 & 99.51 & 56.93\\
				%w/o $\mathcal{L}_{TF}$ & 93.56 &99.89 &59.45 \\\hline
				HCRP (ours) & \textbf{93.86}&\textbf{99.93} &\textbf{62.40} \\
				\bottomrule
			\end{tabular}
		}
		\caption{Accuracy (\%) of 3-way-1-shot scenarios on FewRel 1.0 validation set. Three different settings are designed to illustrate the performance under random, easy, and hard settings.}
		\label{fg-fewrel}
	\end{table}
	This section discusses the effect of hybrid prototype learning.
	As shown in the Table~\ref{ablation}, we conduct an ablation study to verify the effectiveness of hybrid prototypes. Removing local (Model 2) and global prototypes (Model 3) decreases the performance respectively, {\color{black}indicating that both prototypes are essential to represent relations.} Furthermore, we present a 3-way-1-shot task sampled from the FewRel 1.0 validation set, as shown in Table~\ref{case}. %The three relations in support set are ``\textit{mother}'', ``\textit{child}'', and ``\textit{spouse}'' respectively, and the descriptions are also listed. 
	The task can be regarded as a hard task because the three relations are highly similar. Our model correctly classifies the query instance as ``\textit{mother}''.
	We visualize the similarity between the query instance and different relation prototypes, where different columns represent different models. Proto-BERT and HCRP without global prototypes tend to classify the query into the wrong relation ``\textit{child}''. {\color{black}HCRP without local prototypes can correctly predict the relation ``\textit{mother}''. HCRP further correctly predicts with a higher degree of confidence, which proves that hybrid prototypes can better model subtle inter-relation variations for hard tasks.}
	
	\subsubsection{Analysis of Relation-Prototype Contrastive Learning}
\begin{figure}[t!]
		\flushleft
		\includegraphics{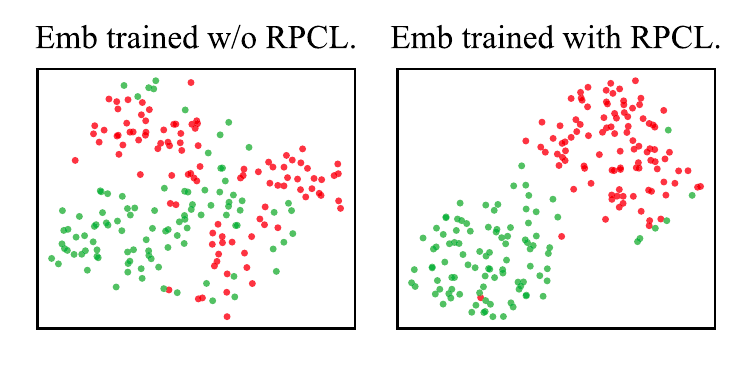}
		\caption{t-SNE plots of instance embeddings trained with or without (w/o) relation-prototype contrastive learning. Two easy-to-confuse relations (``{\em mother}'' and ``{\em child}'') with 100 samples are adopted. Best viewed in color. 
		}
		\label{tsne} 
	\end{figure}

	\begin{table}[t!]
		\centering
		\scalebox{0.8}{
			\begin{tabularx}{\columnwidth}{X}
				\toprule
				Support Set\\	
				\midrule
				\textit{\textbf{mother} | female parent of the subject} \\
				He was the third son of Yang Jian and {\color{red} Dugu Qieluo}, after Yang Yong and {\color{blue} Yang Guang}.\\
				\midrule
				\textit{\textbf{child} | subject has object as biological, foster, and/or adoptive child} \\ 
				He was a son of {\color{blue} Margrethe Rode}, and a brother of writer {\color{red} Helge Rode}. \\
				\midrule
				\textit{\textbf{spouse} | the subject has the object as their spouse (husband, wife, partner, etc.)}\\
				He is the son of Canadian Olympic figure skaters {\color{blue} Don Fraser} and {\color{red} Candace Jones}. \\
				\midrule
				Query Instance\\
				\midrule
				He is the eldest son of actor and director Leo Penn and actress {\color{red} Eileen Ryan}, and the brother of actors Sean Penn and {\color{blue} Chris Penn}.\\
				\multicolumn{1}{c}{$\includegraphics{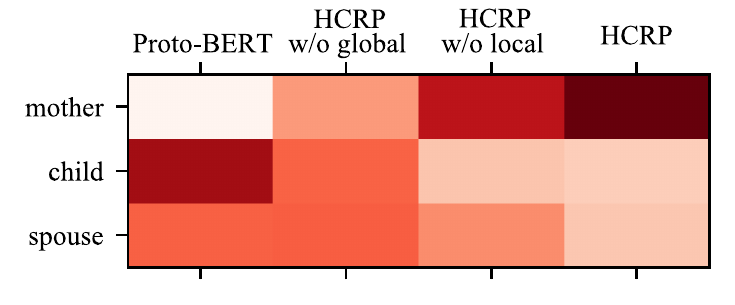}$}\\
				\bottomrule
			\end{tabularx}
		}
		\caption{A real example of 3-way-1-shot hard task. We list the detailed relation names and descriptions, as well as instances. The picture visualizes similarities between the query instance and different prototypes of relations, where different columns represent different models. Best viewed in color. A darker unit indicates a higher value.}
		\label{case}
	\end{table}
	To demonstrate the effectiveness of relation-prototype contrastive learning (RPCL), we first conduct the ablation study, shown in model 4 of Table~\ref{ablation}. It is clear that there is a severe decline in performance if removing the relation-prototype contrastive learning in 5-way-1-shot and 10-way-1-shot settings. {\color{black}As Figure~\ref{tsne} depicts, we visualize the learned embedding spaces with t-SNE \cite{maaten2008visualizing} to intuitively characterize the resulting representations for similar relations. }Specifically, we pick two similar relations ``\textit{mother}'' and ``\textit{child}'' from the FewRel 1.0 validation set, and randomly sample 100 instances for each relation. {\color{black}We can see that embeddings trained with RPCL are clearly separated, which makes classification easier, while those trained without RPCL are lumped together.} By using the relation-prototype contrastive learning, which regards the relation text as anchors and hybrid prototypes as positives and negatives, {\color{black}our model arrives at more discriminative representations, especially for hard tasks.}
	\subsubsection{Analysis of Task-Adaptive Focal Loss}
	\begin{comment}
	\begin{figure}[t!]
	\flushleft
	\includegraphics[width=\linewidth]{Figure/attheatmap.pdf}
	\caption{Similarity visualization between the query instance and different prototypes of relations, where different columns represent different models. The darker units have larger value. The detailed task is shown in Table~\ref{case}.
	}
	\label{attheatmap} 
	\end{figure}
	\end{comment}
	\begin{table*}[t!]
		\centering
		\renewcommand\tabcolsep{5pt}
		\scalebox{0.8}{
			\begin{tabular}{clcl}
				\toprule
				Tasks&Relations & \multicolumn{2}{c}{Weights} \\
				\midrule
				task 1 &performer, director, characters, composer, publisher &0.29 & \multirow{4}*{$\includegraphics[width=0.03\linewidth, height=0.22\columnwidth]{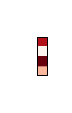}$}\\
				task 2 &performer, has part, location, father, platform, religion &0.19 &\\
				task 3 &located on terrain feature, location of formation, country, work location, location &0.30 & \\
				task 4 &has part, instrument, operating system, military branch, successful candidate &0.22& \\
				\bottomrule
			\end{tabular}
		}
		\caption{An example of a 4-task batch and the task-adaptive weights, where each task has 5 relations ($N$=5).}
		\label{task_weight}
	\end{table*}
	\begin{table}[t!]
		\centering
		%\small
		\renewcommand\tabcolsep{3.6pt}
		\scalebox{0.8}{
			\begin{tabular}{lccc}
				\toprule
				\multirow{2}*{Model} &\multirow{2}*{No.}&5-way&10-way\\ &&1-shot&1-shot\\
				%Model &5-way-1-shot& 10-way-1-shot\\
				\midrule
				HCRP  &1&\textbf{90.90} & \textbf{84.11} \\
				\midrule
				w/o local prototype  &2&88.37 &82.31 \\
				w/o global prototype  &3&86.42 & 77.86\\
				\midrule
				w/o RPCL  &4&87.85 & 79.76\\
				\midrule
				CE loss &5&88.96 &82.75 \\
				CE loss with task weights &6& 89.38 &  83.11\\
				focal loss &7& 89.51 &83.54 \\ 
				\bottomrule
			\end{tabular}
		}
		\caption{Ablation study on FewRel 1.0 validation set showing accuracy (\%).}
		\label{ablation}
	\end{table}
	As shown in Table~\ref{ablation}, we compare our designed task adaptive focal loss with cross entropy (CE) loss (Model 5), cross entropy loss with task weights (Model 6), and focal loss (Model 7). 
	Comparing Model 5 and Model 7, we observe that focal loss achieves higher accuracy than CE loss, and adding the task adaptive weight (Model 1) further improves the performance. In addition, experiments show that the CE loss with task weights also improves performance compared to CE loss. Table~\ref{task_weight} depicts a case study to show the task-adaptive weights. Specifically, we give a sampled mini-batch of tasks from the FewRel 1.0 training set, where the batch size is 4. Each task has 5 relations, which are also listed. The model allocates weights for each task according to the similarity of support set. For example, the relations in task 1 are similar to each other, mainly describing the relations in the art field, so the model assigns a relatively higher task weight. However, the relations in task 2 are very different, so the model allocates a lower weight.
	The ablation experiments and case study prove that our proposed loss can pay more attention to hard tasks in the training process, thus improve the performance. 
	\section{Conclusion}
	This paper focuses on hard few-shot relation extraction tasks and proposes a hybrid contrastive relation-prototype approach. 
	The method proposes a hybrid prototype learning method that generates informative prototypes to model small inter-relation variations. {\color{black}A relation-prototype contrastive learning approach is proposed. Using relation information as anchors, it pulls instances of the same relation class closer in the representation space while pushing dis-similar ones apart. This process further enables the model to acquire more discriminative representations.} In addition, we introduce a task adaptive focal loss to focus more on hard tasks during training to achieve better performance. Experiments have demonstrated the effectiveness of our proposed model. There are multiple avenues for future work. One possible direction is to design a better mechanism for selecting tasks in the training process rather than using random sampling.
	
	\section*{Acknowledgements}
	We would like to thank the anonymous reviewers for their thoughtful and constructive comments. The first author is a visiting student at the StatNLP group of SUTD. This work is supported by the National Key Research and Development Program of China under grant 2018YFB1003804, in part by the National Natural Science Foundation of China under grant 61972043, 61772479, the Fundamental Research Funds for the Central Universities under grant 2020XD-A07-1, the BUPT Excellent Ph.D. Students Foundation under grant CX2020102, and China Scholarship Council Foundation. 
	This research is also supported by Ministry of Education, Singapore, under its Academic Research Fund (AcRF) Tier 2 Programme (MOE AcRF Tier 2 Award No: MOE2017-T2-1-156). Any opinions, findings and conclusions or recommendations expressed in this material are those of the authors and do not reflect the views the Ministry of Education, Singapore.
	
	% Entries for the entire Anthology, followed by custom entries
	\bibliography{anthology,custom}
	\bibliographystyle{acl_natbib}
	
\end{document}